\documentclass[10pt,twocolumn,letterpaper]{article}

\usepackage[pagenumbers]{cvpr} % To force page numbers, e.g. for an arXiv version

\usepackage{graphicx}
\usepackage{amsmath}
\usepackage{amssymb}
\usepackage{booktabs}
\usepackage{algorithm}
\usepackage{algorithmic}
\usepackage{multirow}
\usepackage[accsupp]{axessibility}

\usepackage{colortbl}
\definecolor{Light}{RGB}{247,224,213}
\definecolor{Gray}{gray}{0.90}

\usepackage[pagebackref,breaklinks,colorlinks]{hyperref}

\usepackage[capitalize]{cleveref}
\crefname{section}{Sec.}{Secs.}
\Crefname{section}{Section}{Sections}
\Crefname{table}{Table}{Tables}
\crefname{table}{Tab.}{Tabs.}

\def\cvprPaperID{2488} % *** Enter the CVPR Paper ID here
\def\confName{CVPR}
\def\confYear{2022}

\newcommand{\ParaSpace}{\vspace{-0.3mm}}

\begin{document}

\title{Diversity Matters: Fully Exploiting Depth Clues for Reliable \\ Monocular 3D Object Detection}

\author{Zhuoling Li$^{1}$\thanks{Zhuoling Li and Zhan Qu contributed equally. This work was done when Zhuoling Li was an intern at Huawei Noah’s Ark Lab.}, Zhan Qu$^{2*}$\thanks{Corresponding author.}, Yang Zhou$^{2}$, Jianzhuang Liu$^{2}$, Haoqian Wang$^{1}$, Lihui Jiang$^{2}$ \\
$^{1}$Tsinghua University $^{2}$Huawei Noah’s Ark Lab \\
{\tt\small lzl20@mails.tsinghua.edu.cn \{quzhan, zhouyang116, liu.jianzhuang, jianglihui1\}@huawei.com} \\
{\tt\small wanghaoqian@tsinghua.edu.cn}
}

\maketitle

%%%%%%%%% ABSTRACT
\begin{abstract}
As an inherently ill-posed problem, depth estimation from single images is the most challenging part of monocular 3D object detection (M3OD). Many existing methods rely on preconceived assumptions to bridge the missing spatial information in monocular images, and predict a sole depth value for every object of interest. However, these assumptions do not always hold in practical applications. To tackle this problem, we propose a depth solving system that fully explores the visual clues from the subtasks in M3OD and generates multiple estimations for the depth of each target. Since the depth estimations rely on different assumptions in essence, they present diverse distributions. Even if some assumptions collapse, the estimations established on the remaining assumptions are still reliable. In addition, we develop a depth selection and combination strategy. This strategy is able to remove abnormal estimations caused by collapsed assumptions, and adaptively combine the remaining estimations into a single one. In this way, our depth solving system becomes more precise and robust. Exploiting the clues from multiple subtasks of M3OD and without introducing any extra information, our method surpasses the current best method by more than 20\% relatively on the Moderate level of test split in the KITTI 3D object detection benchmark, while still maintaining real-time efficiency.  
\end{abstract}

%%%%%%%%% BODY TEXT
\section{Introduction}

Significant attention has been drawn by 3D object detection due to its widespread applications in autonomous driving and robotic navigation \cite{arnold2019survey,guo2020deep,grigorescu2020survey,simonelli2020towards}. Inaccurate detection affects the motion planning process directly and could lead to serious accidents. Therefore, the industry has great demand for precise and robust 3D object detection systems. 

Many recently proposed 3D object detection algorithms heavily rely on LiDARs \cite{zheng2021se} and stereo cameras \cite{li2020confidence}, because they are able to perceive the depth information of surroundings directly. Nevertheless, LiDAR sensors are expensive while stereo cameras require exact online calibration \cite{liu2021autoshape}. These limitations make 3D perception using only monocular images promising because it is economical and flexible for deployment.

\begin{figure}[t]
    \centering
    \includegraphics[scale=0.53]{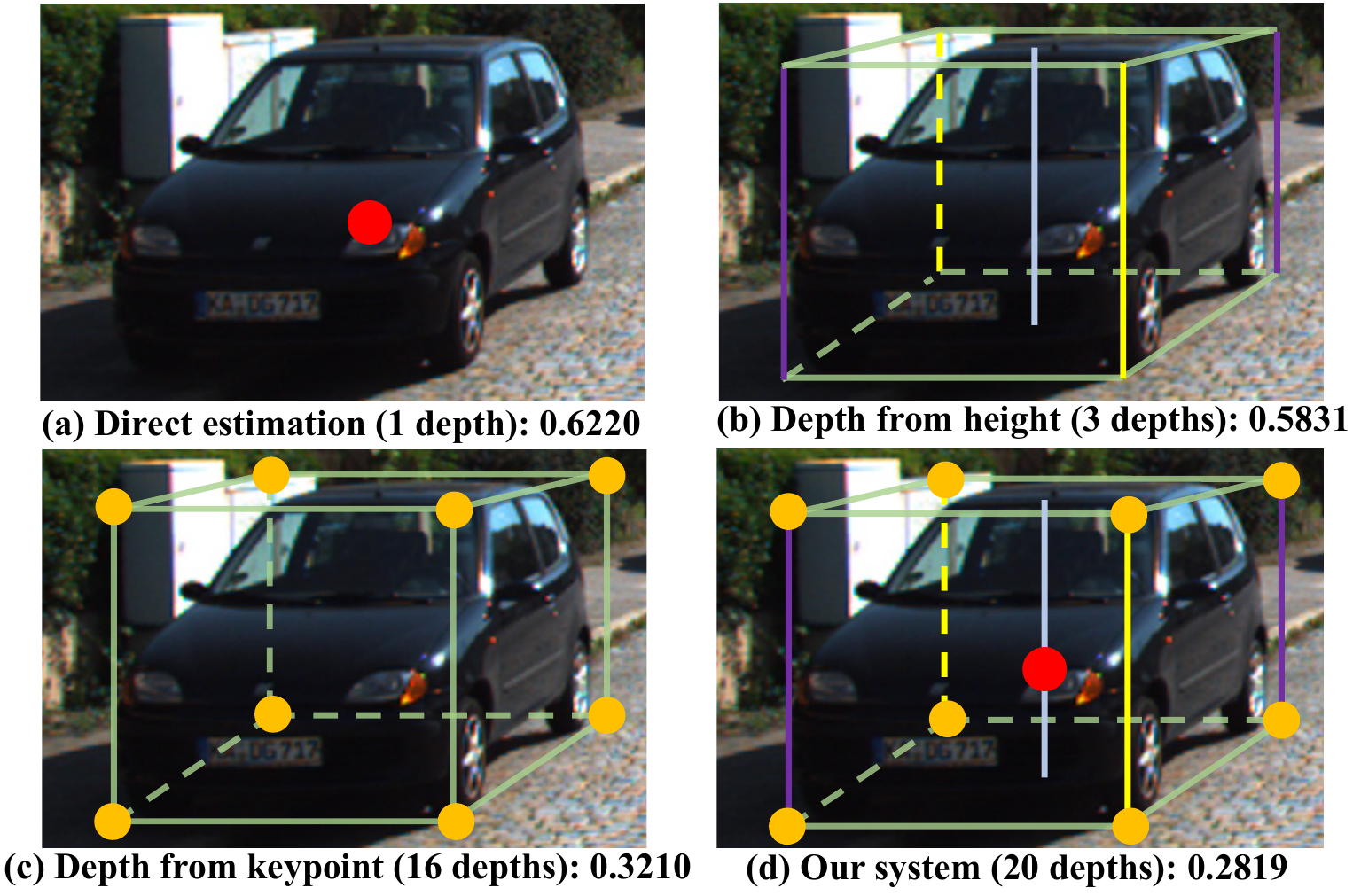}
    \caption{Comparison among various depth solving strategies with different diversity levels. The value below each image is the mean absolute error of depth estimation on the KITTI validation set \cite{geiger2012we} if we always select the most accurate depth from multiple produced estimations. As shown, the error decreases as the diversity of estimations increases.} \label{Idea overview}
    \vspace{-0.4cm}
\end{figure}

The monocular 3D object detection community has achieved prominent progress in recent years. However, there still exists a huge performance gap between the monocular and LiDAR-based methods. This gap is caused by the fact that accurate localization of 3D objects relies on precise depth estimation, and predicting depth from monocular images is an inherently ill-posed problem \cite{ma2020rethinking}, which means the information contained in a single image is insufficient for determining the depths of objects. To compensate the lack of information, current detectors usually resort to some preconceived assumptions. For example, SMOKE \cite{liu2020smoke} assumes that depth can be inferred from visual pixels directly. MonoRCNN \cite{shi2021geometry} hypothesizes the height of a target can be estimated precisely and the camera is an ideal pinhole imaging model \cite{ricolfe2011using}. Nevertheless, these assumptions do not always hold. When the assumptions fail, the single depth produced by a method becomes unreliable.

To address the aforementioned problem, we develop a depth solving system that provides diverse depth estimations for every target. Different from MonoFlex \cite{zhang2021objects} that only utilizes limited information (direct estimation and the heights of objects) and generates similar depths, our method fully exploits various attribute combinations (direct estimation, keypoint, orientation and dimension of an object) to produce 20 depths, which present diverse distributions. Besides, since the 20 depths are separately obtained by solving 20 equations built upon different assumptions, part of the depths are still precise when some assumptions collapse. Figure \ref{Idea overview} illustrates the importance of diversity for monocular depth estimation in the condition that the most accurate one can be selected from predicted depths. When only direct estimation (1 depth) is applied, the mean absolute error (MAE) of depth estimation is 0.6220. In contrast, utilizing our depth solving system, the MAE decreases to 0.2819. 

Although the depths produced by our depth solving system include promising estimations, they also contain outliers. The following problem is how to select promising estimations and combine them into a single value. To that end, we devise a strategy that removes outliers iteratively and integrates the remaining depths based on uncertainty. The experimental results in Section \ref{Study on Depth Selection Strategies} suggest that this strategy is crucial for the overall performance.

Last but not least, considering the uncertainty of both the combined depth and 3D box vertexes, we propose a new scheme, named 3D geometry confidence, to model the conditional 3D confidence. Compared with existing strategies such as modeling the confidence with 3D IOU \cite{chen2021monorun}, our scheme generalizes better.

Incorporating all the techniques, the resulting Monocular 3D detector with diverse depth estimations, named MonoDDE, fully exploits depth clues in monocular images and produce reliable 3D detection boxes in practical applications. Our main contributions are summarized as follows:

$\bullet$$\ $ We point out that the diversity of depth estimation is critical for monocular 3D object detection. Correspondingly, a novel depth solving system that produces 20 depths for every target is developed.

$\bullet$$\ $ We devise a strategy that removes outliers caused by collapsed assumptions and combines the remaining reliable estimations into a single depth. Besides, a new scheme for modeling the conditional 3D confidence is developed. 

$\bullet$$\ $ Using a single model, MonoDDE outperforms the current best method by 20.96\% relatively on the Moderate level of the Car class in KITTI, and ranks 1st and 2nd on the Cyclist and Pedestrian classes, respectively.

\section{Related Work} \label{Related Work}

\noindent \textbf{Monocular 3D object detection.}
According to the form of generated depth, recent monocular 3D object detection algorithms can be mainly categorized into two classes: dense-depth and sparse-depth methods. 

Dense-depth 3D detectors generate depth values for every pixel in an image. The generated dense depth map can be combined with the original RGB image as input to a model for producing 3D object detection boxes \cite{manhardt2019roi,shi2020distance,ma2020rethinking}. Alternatively, it can also be converted to pseudo 3D point clouds firstly and then a LiDAR-based 3D detector is applied on them to derive the results \cite{reading2021categorical,qian2020end,ma2019accurate}. Although the dense-depth methods have achieved impressive results, estimating pixel-wise depths is challenging and requires more complex backbones compared with the strategy that only predicts the depths of several keypoints. This issue has hindered dense-depth methods from further improvement to some extent \cite{zou2021devil}.

Sparse-depth methods only produce one valid depth for every recognized target. Their network structures mostly follow some outstanding 2D detectors, such as Faster RCNN \cite{ren2015faster} and CenterNet \cite{zhou2019objects}. Early sparse-depth methods rely on generating numerous anchors heavily and utilize the information contained in the anchors to regress desired object properties \cite{chen2016monocular,manhardt2019roi,chen20153d}. However, the anchor-generating process introduces non-negligible noise and increases computation burden \cite{liu2020smoke}. Recent sparse-depth 3D detectors are mainly center-based \cite{liu2021autoshape,zhou2021monocular}, which represent objects by their 2D centers \cite{liu2020smoke} or projected 3D centers \cite{ma2021delving}. This anchor-free structure has led to simpler model structures, fewer hyper-parameters and better detection precision \cite{li2020rtm3d}. Our proposed MonoDDE is also center-based.

\vspace{1mm}
\noindent \textbf{Sparse depth estimation.} Experimental results in previous works have shown that depth estimation is the most crucial step in center-based methods \cite{zhang2021objects}, and existing sparse depth estimation can be roughly divided into 3 strategies, \textit{direct depth estimation} \cite{liu2020smoke}, \textit{depth from height} \cite{shi2021geometry} and \textit{pespective-n-point} (PnP) \cite{li2021monocular}.

Among the three strategies, \textit{direct depth estimation} is the easiest for implementation. Taking monocular images as input, it completely relies on a deep neural network to explore visual clues and infer depths \cite{liu2020smoke,zhou2021monocular}. Besides, since \textit{direct depth estimation} does not require manual annotation, its precision can be improved conveniently via large-scale self-supervised pre-training without labels \cite{park2021pseudo}. Nevertheless, since monocular depth estimation is an ill-posed problem, the estimated values are not reliable when there exists a significant domain gap between training and testing images.

\textit{Depth from height} computes depths based on the pixel heights and estimated physical heights of targets \cite{shi2021geometry}. Since the physical heights of objects belonging to the same category are similar, \textit{depth from height} generalizes better than \textit{direct depth estimation} \cite{liu2021autoshape}. However, estimating physical heights is still an ill-posed problem.

In contrast to \textit{direct depth estimation} and \textit{depth from height}, PnP incorporates all the dimension, orientation and keypoint information of an object to construct geometric constraints \cite{li2021monocular,li2020rtm3d,liu2021autoshape} and uses the \textit{least squares method} \cite{menke2015review} to obtain its location. Therefore, PnP exploits the information more efficiently. However, all the equations in PnP are closely coupled with each other \cite{liu2021autoshape}. This issue causes the difficulty to model the uncertainty of every depth individually.

\section{Preliminary}
To present our method clearly, we first review the target of monocular 3D object detection. Afterwards, the mathematical forms of the three depth estimation strategies mentioned in Section \ref{Related Work} are given, which are \textit{direct depth estimation}, \textit{depth from height} and PnP.

\ParaSpace
\subsection{Monocular 3D Object Detection}
\ParaSpace
Given a single image, monocular 3D object detection aims to find every object of interest, identify its category and estimate a 3D box $B$ that contains the object properly. The 3D box $B$ can be further divided into 3 properties, i.e., the 3D center location $(x, y, z)$, dimension $(h, w, l)$ and orientation (yaw angle) $\theta$. The roll and pitch angles of objects are set to 0 following the KITTI \cite{geiger2012we} setting.

Among these properties, the dimension and orientation are strongly related to the visual appearance and can be learned by a network \cite{karpathy2014large}, while the 3D location is challenging to obtain. This is because producing an accurate 3D location is built upon the premise of precise depth estimation. Thus, how to estimate the depth correctly is the most important research topic in monocular 3D object detection.

\ParaSpace
\subsection{Depth Estimation Strategies} \label{Estimate 3D location}
\ParaSpace

\noindent \textbf{Direct depth estimation.} Given an input image $I$, \textit{direct depth estimation} relies on the appearance of an object and its surrounding pixels to regress depth $z$ directly. Afterwards, utilizing the projected 3D center estimation $(u_{c}, v_{c})$, $x$ and $y$ are determined as:
\begin{align} 
x = \frac{(u_{c} - c_{u})z}{f_{x}}, \quad y = \frac{(v_{c} - c_{v})z}{f_{y}}, \label{Eq1}
\end{align}
where $(c_{u}, c_{v})$ represents the coordinate of the principle point, and $f_{x}$ and $f_{y}$ are the horizontal and vertical focal lengths, respectively.

\vspace{1mm}
\noindent \textbf{Depth from height.} The \textit{depth from height} strategy tackles depth estimation by decoupling it as predicting the physical height $h$ and pixel height $h'$ of an object. The process of computing $z$ given $h$ and $h'$ can be formulated as:
\begin{align}
z = \frac{f_{y} h}{h'}. \label{Eq3}
\end{align}
After obtaining $z$, $x$ and $y$ are calculated using Eq.~(\ref{Eq1}).

% Instead of regressing $z$ directly, \textit{Depth from height} computes depth through producing two variables ($h$ and $h'$) more robust for estimation. Therefore, it often shows better generalization ability compared with \textit{Direct depth estimation} \cite{liu2021autoshape}. However, estimating $h$ is actually also an ill-posed problem, which restricts the final detection accuracy.

\vspace{1mm}
\noindent \textbf{Pespective-n-point.} Since objects in 3D object detection are represented as cuboids, we can use their geometric constraints to obtain their 3D locations based on the \textit{least squares method}.

Denoting the position of a 3D keypoint under the object coordinate system as $\mathbf{P}^{o}=(x^{o}, y^{o}, z^{o})^{T}$, it can be transformed to the camera coordinate system with respect to the rotation matrix $\mathbf{R}$ and translation vector $\mathbf{T}$ as:
\begin{align}
[x^{c}, y^{c}, z^{c}]^{T} = \mathbf{R} [x^{o}, y^{o}, z^{o}]^{T} + \mathbf{T},   \label{Eq4}
\end{align}
where $\mathbf{P}^{c}=(x^{c}, y^{c}, z^{c})^{T}$ represents the location of this 3D point under the camera coordinate system, and
\begin{align}
\mathbf{R} = \begin{bmatrix} cos\theta & 0 & sin\theta \\ 0 & 1 & 0 \\ -sin\theta & 0 & cos\theta \end{bmatrix}, \quad  
\mathbf{T} = \begin{bmatrix} x, & y, & z \end{bmatrix}^{T}. \label{Eq5}
\end{align}

Afterwards, given the camera intrinsic matrix $\mathbf{K}$, we can project $\mathbf{P}^{c}$ to a point in the 2D pixel coordinate system as $(u, v)$:
\begin{align}
& \lambda [u, v, 1]^{T} = \mathbf{K}[x^{c}, y^{c}, z^{c}]^{T}, \label{Eq7} \\
& \mathbf{K} = \begin{bmatrix} f_{x} & 0 & c_{u} \\ 0 & f_{y} & c_{v} \\ 0 & 0 & 1 \end{bmatrix}, \quad \lambda = z^{c}. \label{Eq8}
\end{align}

\begin{figure*}[htbp]
    \centering
    \includegraphics[scale=0.64]{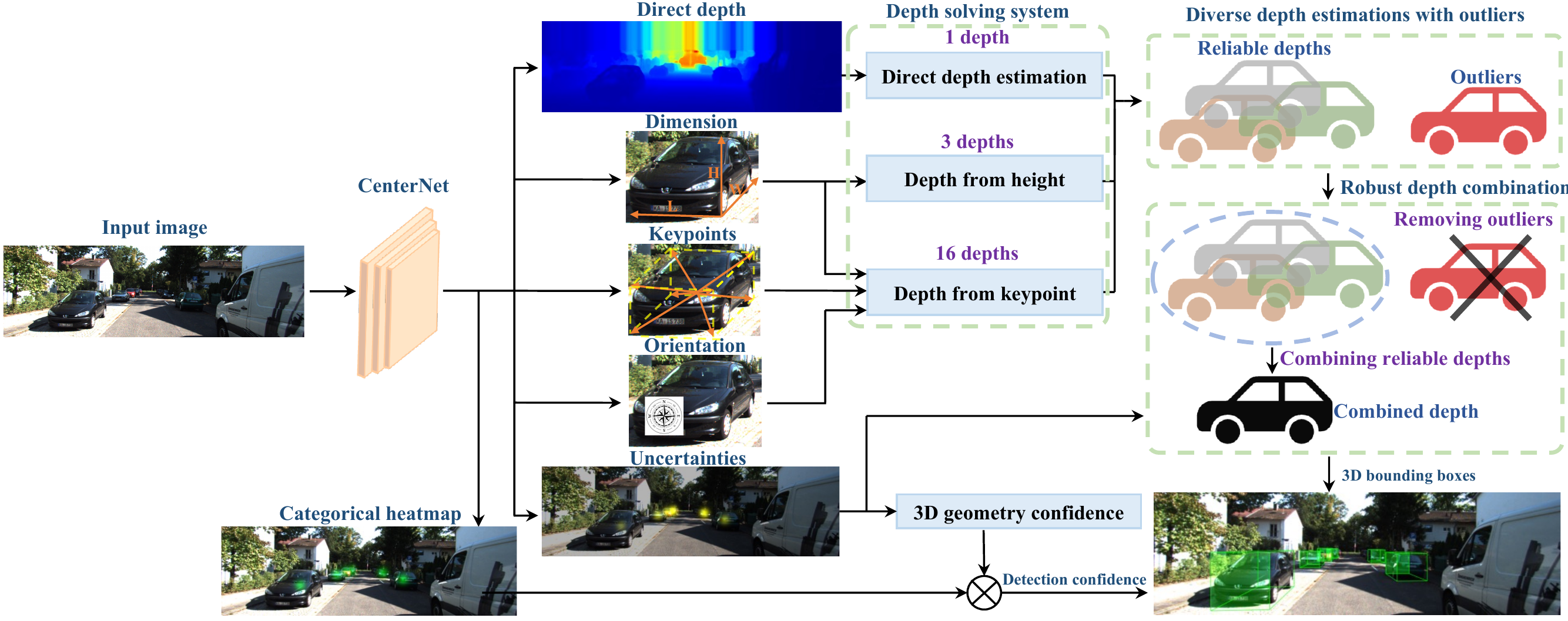}
    \caption{Overall pipeline of MonoDDE.} \label{Overall pipeline}
    \vspace{-0.4cm}
\end{figure*}

Hereby, the geometric relations between any point in the object coordinate system and its corresponding pixel on the 2D imaging plane are described by Eqs.~(\ref{Eq4})--(\ref{Eq8}). In these relations, $\mathbf{P}^{o}$ is pre-defined manually, $\mathbf{K}$ is known, and $\mathbf{R}$ and $(u, v)$ are estimated by a network. Thus, $\mathbf{T} = [x, y, z]^{T}$ contains the only variables waiting to be computed. Since every 3D keypoint provides 2 geometric constraints, we can obtain $x$, $y$ and $z$ simultaneously using the \textit{least squares method} if we have at least 2 keypoints.

% In contrast to \textit{Direct depth estimation} and \textit{Depth from height} that only use limited information, \textit{Depth from keypoint} incorporates all the dimension, orientation and keypoint information of objects to determine their locations. Nevertheless, \textit{Depth from keypoint} has two serious limitations. First of all, how to estimate the depth $z$ precisely is the mostly concerned issue in monocular 3d object detection. Calculating $x$, $y$ and $z$ simultaneously distracts this solving process. Secondly, all equations in \textit{Depth from keypoint} are closely coupled with each other, and it is difficult to model the uncertainty of each equation separately.

\section{Method}
This section details our proposed method and how MonoDDE is implemented. 
%In this Section, we introduce how MonoDDE is implemented. First of all, Section \ref{Overall framework} introduces the overall framework. Afterwards, Section \ref{Diverse depth estimations} and \ref{Reliable depth selection} explain how we produce multiple diverse depth estimations and select out reliable ones. Finally, Section \ref{Detection heads} presents the implementation details of adopted network heads and their corresponding loss functions.

\ParaSpace
\subsection{Overall Framework} \label{Overall framework}
\ParaSpace
The overall framework of MonoDDE is illustrated in Figure \ref{Overall pipeline}. MonoDDE employs CenterNet \cite{zhou2019objects} as the base model for producing discriminative representation. Specifically, for any input image $I$, DLA34 \cite{yu2018deep} is adopted as the backbone of CenterNet for extracting features. We establish several network heads to regress object properties, including categorical heatmap, 2D bounding box, dimension, keypoint offsets, orientation, depth and multiple uncertainty items. Based on the regressed properties, our proposed depth solving system produces 20 diverse depths in different ways. Subsequently, the developed robust depth combination module filters out outlier values and combines the remaining estimations as a single depth. Taking this depth value into Eq.~(\ref{Eq1}), we get the location of the target and further its 3D box with the regressed dimension and orientation. In addition, the detection confidence obtained by our 3D geometry confidence (Section \ref{Mixed determinacy confidence}) is responsible for modeling the probability that a target is recognized correctly.

\ParaSpace
\subsection{Diverse Depth Estimations} \label{Diverse depth estimations}
\ParaSpace
In this work, we expect our developed depth solving system to possess three key characteristics: (1) It should concentrate on obtaining depth $z$ rather than computing $x$, $y$ and $z$ together. (2) In contrast to existing methods, it should produce multiple and diverse estimation values. (3) It should make full use of all available information, including visual clue, estimated target center, dimension, orientation and keypoints.

To realize the above goal, we first revisit the geometric constraints described in Section \ref{Estimate 3D location}. Combining Eqs.~(\ref{Eq4})--(\ref{Eq8}), we can simplify the relation between a 3D keypoint under the object coordinate system $P^{o}=(x^{o}, y^{o}, z^{o})$ and its corresponding pixel $(u, v)$ as:
\begin{align}
\begin{bmatrix} -1 & 0 & \tilde{u} \\ 0 & -1 & \tilde{v} \end{bmatrix} \begin{bmatrix} x \\ y \\ z \end{bmatrix} = \begin{bmatrix} \tilde{u} \\ \tilde{v} \end{bmatrix} \mathbf{A} + \mathbf{B}, \label{Eq9}
\end{align}
where
\begin{align}
& \tilde{u} = \frac{u - c_{u}}{f_{x}} \label{Eq10}, \quad
\tilde{v} = \frac{v - c_{v}}{f_{y}}, \\
& \mathbf{A} = \begin{bmatrix} x^{o} sin\theta - z^{o} cos\theta \end{bmatrix} \label{Eq12}, \\
& \mathbf{B} = \begin{bmatrix} x^{o} cos\theta + z^{o} sin\theta \\ y^{o} \end{bmatrix}. \label{Eq13}
\end{align}

It can be observed from Eq.~(\ref{Eq9}) that $x$, $y$ and $z$ appear in the same equation, which hinders this system from only obtaining $z$. In order to solve this problem, we need to resort to some extra prior knowledge.

Through experiments, we observe that most centers of objects can be recognized precisely. More than 85\% of estimated object centers fall within 1 pixel around their corresponding ground truth points. Hence, Eq. (\ref{Eq1}) can be used as the prior. By inserting Eq.~(\ref{Eq1}) into Eq.~(\ref{Eq9}), Eq.~(\ref{Eq9}) can be reformulated as:
\begin{align}
(\tilde{u} - \tilde{u}_{c}) z &= \mathbf{A} \tilde{u} + x^{o} \cos\theta + z^{o} \sin\theta, \label{Eq14} \\ 
(\tilde{v} - \tilde{v}_{c}) z &= \mathbf{A} \tilde{v} + y^{o}, \label{Eq15}
\end{align}
where $\tilde{u}_{c} = \frac{u_{c} - c_{u}}{f_{x}}$ and $\tilde{v}_{c} = \frac{v_{c} - c_{v}}{f_{y}}$.

\begin{figure}[ht]
    \vspace{-0.2cm}
    \centering
    \includegraphics[scale=0.57]{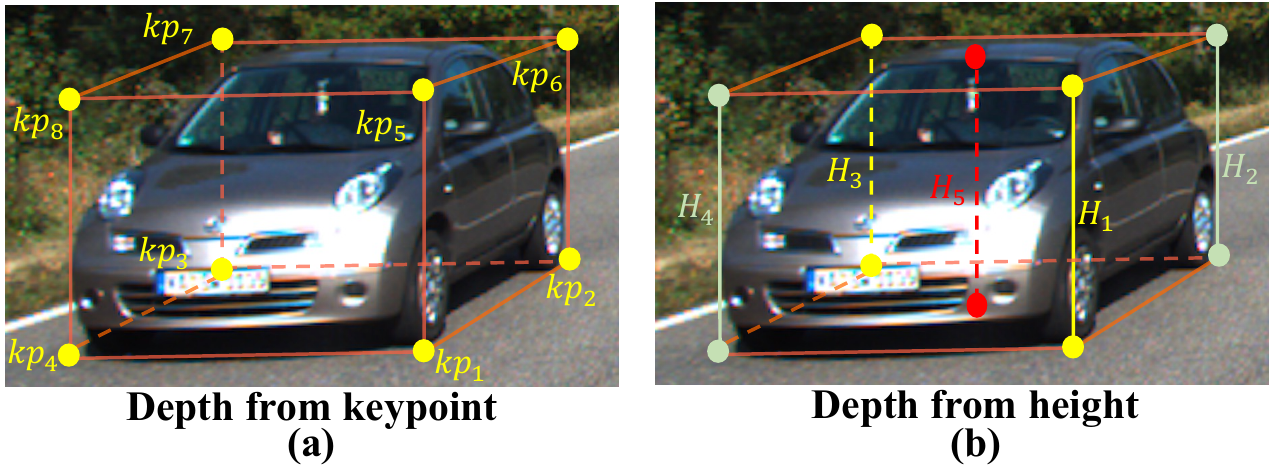}
    \caption{Depths from keypoints and heights.} \label{Depth solving system}
    \vspace{-0.4cm}
\end{figure}

In this way, Eq.~(\ref{Eq9}) is decoupled into two independent equations, Eqs.~(\ref{Eq14}) and (\ref{Eq15}), which focus on solving for $z$. The geometric relation between every 3D vertex and its corresponding projected pixel can result in 2 separate depths. In our implementation, as shown in Figure \ref{Depth solving system} (a), we select 8 vertexes of a 3D box as the keypoints to calculate the depths, which provide 16 diverse estimation values.

Furthermore, we incorporate the \textit{direct depth estimation} and \textit{depth from height} strategies into the depth solving system. Specifically, \textit{direct depth estimation} regresses 1 depth value of the projected 3D center like \cite{liu2020smoke}. For \textit{depth from height}, as shown in Figure \ref{Depth solving system} (b), we split the heights of the center vertical line and corner vertical lines into three groups, \{$H_{5}$\}, \{$H_{1}$, $H_{3}$\}, and \{$H_{2}$, $H_{4}$\}, which is similar to \cite{zhang2021objects}. The depth of an object can be obtained using the center vertical line $H_{5}$ and Eq.~(\ref{Eq3}) or by averaging the depths generated using the opposite corner vertical lines (\{$H_{1}$ and $H_{3}$\} or \{$H_{2}$ and $H_{4}$\}).

Hereby, we have established a depth solving system that can output 20 diverse depths, 16 from our newly proposed geometric constraints (\textit{depth from keypoint}), 1 from \textit{direct depth estimation}, and 3 from \textit{depth from height}. The following problem is how to select reliable depths from them.

\ParaSpace
\subsection{Robust Depth Combination} \label{Robust depth combination}
\ParaSpace
In this subsection, we present the strategies for selecting and combining promising depths.

\vspace{1mm}
\noindent \textbf{Output distribution.} Assuming each estimated depth follows the Gaussian distribution \cite{lu2021geometry}, the model learns to predict the mean and variance of this distribution by minimizing:
\begin{align}
L_{\sigma} = \frac{|p - p^{*}|}{\sigma} + log\sigma, \label{Eq16}
\end{align}
where $p$ and $\sigma$ are the predicted mean and standard deviation of the output distribution, respectively, and $p^{*}$ represents the ground truth. Note that $\sigma$ is learned implicitly from Eq. (\ref{Eq16}) without the need of ground truth. More details are given in \cite{kendall2017uncertainties, choi2019gaussian} about why the distribution can be captured by the network in this way.

Moreover, we define the distribution of a set $S = \{s_{i}\}_{i=1}^{N}$, which contains $N$ Gaussian distribution variables $s_{i} \sim \mathcal{N}(\mu_{i}, \sigma^{2}_{i})$, as a new Gaussian distribution, because all the heads predict the depths of the same target. It is the weighted sum of $\{s_{i}\}_{i=1}^{N}$ and the weights $\{\omega_{i}\}_{i=1}^{N}$ are derived via the \textit{weighted least squares method} \cite{asparouhov2010weighted}:

\begin{align}
\omega_{i} = \frac{ 1 / \sigma^{2}_{i}} {\sum\limits_{j=1}^{N} 1 / \sigma^{2}_{j} }. \label{Eq17}
\end{align}
Hence, the mean $\mu_{s}$ and variance $\sigma^{2}_{s}$ of $S$ are calculated as:
\begin{align}
\mu_{s} = \sum\limits_{i=1}^{N} \omega_{i} \mu_{i}, \  % \label{Eq18} \\
\sigma^{2}_{s} = \sum\limits_{i=1}^{N} \omega_{i}^{2} \sigma^{2}_{i} \label{Eq19}.
\end{align}

\vspace{1mm}
\noindent \textbf{Selecting and combining reliable depths.} We first train our model to predict the means and variances of the 20 depth distributions using Eq.~(\ref{Eq16}), and compose the 20 distributions as the set $S$. Since both $S$ and its contained variables $\{s_{i}\}_{i=1}^{N}$ are treated as Gaussian distributions, we can filter out outliers based on the 3$\sigma$ rule \cite{pukelsheim1994three}, and devise a robust algorithm similar to the expectation-maximization (EM) algorithm \cite{dempster1977maximum}.

\begin{algorithm}[htb]
\caption{Robust depth selection and combination.} \label{Robust depth combination pseudo code}
\begin{algorithmic}[1]
    \REQUIRE Estimated depths $\{z_{i}\}_{i=1}^{N}$ and their corresponding variances $\{\sigma^{2}_{i}\}_{i=1}^{N}$
    \STATE Initialize an empty set $S=\varnothing$
    \STATE Put $z_{k}$ to $S$ if $\sigma^{2}_{k} = \min\{\sigma^{2}_{1}, \sigma^{2}_{2}, ..., \sigma^{2}_{N}\}$
    \WHILE{True}
    \STATE Update $\mu_{s}$ and $\sigma^{2}_{s}$ according to Eqs.~(\ref{Eq17})--(\ref{Eq19})
    \STATE $S_{new}=\varnothing$
    \FOR{$i=1$ to $N$}
    \STATE $S_{new} \leftarrow S_{new} \cup \{z_{i}\}$ if $z_{i} \in ( \mu_{s}-3\sigma_{s}, \mu_{s}+3\sigma_{s})$ and $z_{i} \notin S$.
    \ENDFOR
    \IF{$S_{new}$ is $\varnothing$}
    \STATE break
    \ENDIF
    \STATE $S \leftarrow S \cup S_{new}$
    \ENDWHILE
    \ENSURE $\mu_{s}$
\end{algorithmic}
\end{algorithm}

In this algorithm, we first initialize $S$ as an empty set and put the depth with minimum variance to $S$. For the maximization step, $\mu_{s}$ and $\sigma^{2}_{s}$ are updated using Eqs.~(\ref{Eq17})--(\ref{Eq19}). During the expectation step, the depths that fall into $(\mu_{s}-3\sigma_{s}, \mu_{s}+3\sigma_{s})$ are added to $S$. We repeat the maximization and expectation steps until $\mu_{s}$ and $\sigma^{2}_{s}$ converge. Afterwards, all depths falling out of $(\mu_{s}-3\sigma_{s}, \mu_{s}+3\sigma_{s})$ are regarded as outliers and removed.

In this way, the reliable depths are contained in $S$. We directly employ the final $\mu_{s}$ as the combined depth $z_{c}$ for subsequent operations. The pseudo code of the robust depth combination is given in Algorithm \ref{Robust depth combination pseudo code}.

\ParaSpace
\subsection{3D Geometry Confidence} \label{Mixed determinacy confidence}
\ParaSpace
Let ${\rm P}_{m}$ be the probability (also called confidence) that a target is detected correctly. Following \cite{li2020rtm3d} with the probability chain rule, it is factorized into two items:
\begin{align}
{\rm P}_{m} = {\rm P}_{3d|2d} \cdot {\rm P}_{2d}, \label{Eq20}
\end{align}
where ${\rm P}_{2d}$ is represented by the categorical heatmap score and ${\rm P}_{3d|2d}$ denotes the conditional 3D confidence. Previous methods often model ${\rm P}_{3d|2d}$ with 3D IOU \cite{chen2021monorun,simonelli2021we,wang2022probabilistic}. However, since the training images are used to train the model and the validation images are unseen, the mean 3D box IOU of the model on the training images is significantly higher than that on the validation images. Due to the large IOU gap, directly employing 3D IOU in the training stage to train the network and regarding the predicted IOU as ${\rm P}_{3d|2d}$ lead to poor results in the validation stage. Meanwhile, some works have indicated that models trained with implicit supervision generalize better \cite{zhao2020makes}. Hence, we model ${\rm P}_{3d|2d}$ based on the estimated variance in Eq.~(\ref{Eq16}), which is implicitly learned. Specifically, following \cite{zhang2021objects}, we define the confidence $d$ of an estimation item with respect to its  variance $\sigma^{2}$ as:
\begin{align}
d = 1 - \min \{ \sigma^{2}, 1 \}. \label{Eq21}
\end{align}

In this work, we model ${\rm P}_{3d|2d}$ as the weighted sum of two items, the combined depth confidence $d_{c}$ and the 3D box confidence $d_{b}$:
\begin{align}
{\rm P}_{3d|2d} = \omega_{c} d_{c} + \omega_{b} d_{b}, \label{Eq23}
\end{align}
where $\omega_{c}$ and $\omega_{b}$ are calculated based on $\sigma_{c}^{2}$ and $\sigma_{b}^{2}$ using Eq.~(\ref{Eq17}). ${\rm P}_{3d|2d}$ in Eq. (\ref{Eq23}) is our devised 3D geometry confidence.

The combined depth variance $\sigma_{c}^{2}$ for determining $d_{c}$ is learned with Eq.~(\ref{Eq16}). We do not directly use $\sigma_{s}^{2}$ as $\sigma_{c}^{2}$ because we observe that the estimated $\sigma_{c}^{2}$ leads to a more precise value. Meanwhile, similar to Eq. (\ref{Eq16}), the variance of the 3D box $\sigma_{b}^{2}$ is obtained via minimizing:
\begin{align}
L_{b} = \frac{ \sum\limits_{i=1}^{8} |v_{i} - v_{i}^{*}|} {\sigma_{b}} + log\sigma_{b}, \label{Eq22}
\end{align}
where $\{v_{i}\}_{i=1}^{8}$ denote the coordinates of the 8 3D box vertexes and $\{v_{i}^{*}\}_{i=1}^{8}$ are their ground truth.

\ParaSpace
\subsection{Network Heads} \label{Detection heads}
\ParaSpace
This subsection describes the implementation of the detection heads briefly. Each head comprises two convolutional layers and one batch normalization layer.

\vspace{1mm}
\noindent \textbf{Categorical heatmap.} It is responsible for distinguishing the categories of objects and localizing target points. In this work, we employ projected 3D centers as the ground truth of the target points, and the representation decoupling strategy devised in MonoFlex \cite{zhang2021objects} is adopted to tackle truncated objects. The loss function follows \cite{liu2020smoke}.

\vspace{1mm}
\noindent \textbf{Orientation.} Similar to \cite{mousavian20173d}, we regress the observation angle $\alpha$ instead of the yaw angle $\theta$, and train the network with the MultiBin loss. $\alpha$ is split into 4 bins like \cite{brazil2020kinematic}, and then $\theta$ is obtained based on $\alpha$.

\vspace{1mm}
\noindent \textbf{Dimension.} To be consistent with existing works, we predict the log-scale offsets of dimensions rather than directly outputting absolute sizes. Refer to \cite{zhou2019objects} for details.

\vspace{1mm}
\noindent \textbf{Keypoints.} Following \cite{zhang2021objects}, MonoDDE regresses the offsets from target points to 10 pre-defined 2D keypoints, which include 8 vertexes, the bottom center and top center of a 3D bounding box.

\vspace{1mm}
\noindent \textbf{Depth.} This head is responsible for producing the direct estimation depth $z$. Notably, instead of estimating the absolute value of $z$ directly, MonoDDE learns to fit its exponentially transformed form in \cite{chen2020monopair}.

\vspace{1mm}
\noindent \textbf{Uncertainty.} Based on Eq.~(\ref{Eq16}), we enforce the network to capture the uncertainties (variances) of the 20 depth values, the combined depth $z_{c}$, and the 3D box.

\section{Experiments}
%In this Section, we conduct extensive experiments to verify the detection capability of MonoDDE. First of all, Section \ref{Experimental Details} introduces the evaluation benchmark and training details. Secondly, Section \ref{Quantitative Results} contrasts the performance of MonoDDE with existing methods. Afterwards, Section \ref{Study on Depth Estimation Methods} and \ref{Study on Depth Selection Strategies} analyze the effectiveness of our proposed depth solving system and depth selection strategies, respectively. Finally, Section \ref{Qualitative Results} further proves the superiority of MonoDDE through visualizing some detection results.

\begin{table*}[t] 
    \vspace{-0.2cm}
    \centering
    \resizebox{164mm}{32mm}{
    \begin{tabular}{c|c|c|ccc|ccc|ccc|c}
    \hline \hline
    \multirow{2}{*}{Method} & \multirow{2}{*}{Depth} & \multirow{2}{*}{Extra} & \multicolumn{3}{c|}{Test, ${\rm AP}_{3D}70$ (\%)} & \multicolumn{3}{c|}{Test, ${\rm AP}_{BEV}70$ (\%)} & \multicolumn{3}{c|}{Val, ${\rm AP}_{3D}70$ (\%)} & \multirow{2}{*}{Time (s)} \\
    \cline{4-12}
    & & & Easy & Moderate & Hard & Easy & Moderate & Hard & Easy & Moderate & Hard & \\
    \cline{1-13}
    M3D-RPN \cite{brazil2019m3d} & E & - & 14.76 & 9.71 & 7.42 & 21.02 & 13.67 & 10.23 & 14.53 & 11.07 & 8.65 & 0.16\\
    SMOKE \cite{liu2020smoke} & E & - & 14.03 & 9.76 & 7.84 & 20.83 & 14.49 & 12.75 & - & - & - & 0.03 \\
    MonoPair \cite{chen2020monopair} & E & - & 13.04 & 9.99 & 8.65 & 19.28 & 14.83 & 12.89 & 16.28 & 12.30 & 10.42 & 0.06 \\
    Monodle \cite{ma2021delving} & E & - & 17.23 & 12.26 & 10.29 & 24.79 & 18.89 & 16.00 & 17.45 & 13.66 & 11.68 & 0.04 \\
    GrooMeD-NMS \cite{kumar2021groomed} & E & - & 18.10 & 12.32 & 9.65 & 26.19 & 18.27 & 14.05 & 19.67 & 14.32 & 11.27 & 0.12 \\
    Kinematic3D \cite{brazil2020kinematic} & E & Video & 19.07 & 12.72 & 9.17 & 26.69 & 17.52 & 13.10 & 19.76 & 14.10 & 10.47 & 0.12 \\
    CaDDN \cite{reading2021categorical} & E & Depth & 19.17 & 13.41 & 11.46 & 27.94 & 18.91 & 17.19 & 23.57 & 16.31 & 13.84 & 0.63 \\
    DFR-Net \cite{zou2021devil} & E & Depth & 19.40 & 13.63 & 10.35 & 28.17 & 19.17 & 14.84 & \underline{24.81} & \underline{17.78} & 14.41 & 0.18 \\
    MonoEF \cite{zhou2021monocular} & E & - & 21.29 & 13.87 & 11.71 & 29.03 & 19.70 & \underline{17.26} & - & - & - & 0.03 \\
    \cline{1-13}
    MonoRCNN \cite{shi2021geometry} & H & - & 18.36 & 12.65 & 10.03 & 25.48  & 18.11 & 14.10  & 16.61  & 13.19 & 10.65 & 0.07 \\
    \cline{1-13}
    RTM3D \cite{li2020rtm3d} & P & - & 14.41 & 10.34 & 8.77 & 19.17 & 14.20 & 11.99 & - & - & - & 0.05 \\
    KM3D \cite{li2021monocular} & P & - & 16.73 & 11.45 & 9.92 & 23.44 & 16.20 & 14.47 & - & - & - & 0.03 \\
    Autoshape \cite{liu2021autoshape} & P & CAD & \underline{22.47} & \underline{14.17} & 11.36 & \underline{30.66} & \underline{20.08} & 15.95 & 20.09 & 14.65 & 12.07 & 0.04 \\
    \cline{1-13}
    MonoFlex \cite{zhang2021objects} & EH & - & 19.94 & 13.89 & \underline{12.07} & 28.23 & 19.75 & 16.89 & 23.64 & 17.51 & \underline{14.83} & 0.03 \\

    \cline{1-13}
    MonoDDE (ours) & EHK & - & \textbf{24.93} & \textbf{17.14} & \textbf{15.10} & \textbf{33.58} & \textbf{23.46} & \textbf{20.37} & \textbf{26.66} & \textbf{19.75} & \textbf{16.72} & 0.04 \\
    \hline \hline
    \end{tabular}}
    \caption{Performance comparison between MonoDDE and recent SOTAs on the Car class of KITTI. They are sorted according to their depth solving strategies shown in the 2nd column (E: Direct depth estimation, H: Depth from height, P: PnP, and K: Depth from keypoint).} \label{Performance contrast with SOTAs on Car}
    \vspace{-0.3cm}
\end{table*}

\noindent \textbf{Dataset.} Our method is evaluated on the KITTI 3D object detection benchmark \cite{geiger2012we}, which comprises 7481 images for training and 7518 images for testing. Since the annotations of the testing data are not available, following \cite{zhou2018voxelnet}, we further divide the training data into the training set (3712 images) and validation set (3769 images). Our reported detection classes include Car, Pedestrian and Cyclist. Besides, the objects in KITTI have been categorized into three difficulty levels (Easy, Moderate and Hard) according to their pixel heights, occlusion ratios, etc.

\noindent \textbf{Evaluation metrics.} The average precision (AP) of 3D bounding boxes and bird's-eye view (BEV) map are main metrics for comparing performance. Following \cite{simonelli2019disentangling}, 40 recall positions are sampled to calculate AP. The IOU thresholds are 0.7 for Car, and 0.5 for Pedestrian and Cyclist.

\vspace{1mm}
\noindent \textbf{Implementation details.} MonoDDE is trained for 100 epochs with the initial learning rate 3e-4. The weights of the model are updated using the AdamW optimizer \cite{loshchilov2017decoupled} and the learning rate is decayed at the 80th and 90th epochs \cite{qu2021focus}. The batch size is set to 8 and the whole training process is conducted on a single Tesla V100 GPU. Random horizontal flipping is the only augmentation operation.

\ParaSpace
\subsection{Quantitative Results} \label{Quantitative Results}
\ParaSpace
We compare our method with recent SOTA counterparts of monocular 3D object detection on the KITTI benchmark. The detection results of the Car category are reported in Table \ref{Performance contrast with SOTAs on Car}, and the comparison on Pedestrian and Cyclist is given in Table \ref{Performance contrast on Pedestrian and Cyclist}. For the convenience of observation, the best and second-best results are in bold and underlined, respectively.

\begin{table}[htbp] 
    \vspace{-0.2cm}
    \centering
    \resizebox{82mm}{16mm}{
    \begin{tabular}{c|ccc|ccc}
    \hline \hline
    \multirow{3}{*}{Method} & \multicolumn{6}{c}{Test, ${\rm AP}_{3D}50$ (\%)} \\
    \cline{2-7}
    & \multicolumn{3}{c|}{Pedestrian} & \multicolumn{3}{c}{Cyclist} \\
    \cline{2-7}
    & Easy & Moderate & Hard & Easy & Moderate & Hard \\
    \cline{1-7}
    M3D-RPN \cite{brazil2019m3d} & 4.92 & 3.48 & 2.94 & 0.94 & 0.65 & 0.47 \\
    MonoPair \cite{chen2020monopair} & 10.02 & 6.68 & 5.53 & 3.79 & 2.12 & 1.83 \\ 
    CaDDN \cite{reading2021categorical} & \textbf{12.87} & \textbf{8.14} & \textbf{6.76} & \textbf{7.00} & 3.41 & \underline{3.30} \\
    DFR-Net \cite{zou2021devil} & 6.09 & 3.62 & 3.39 & 5.69 & \underline{3.58} & 3.10 \\
    MonoFlex \cite{zhang2021objects} & 9.43 & 6.31 & 5.26 & 4.17 & 2.35 & 2.04 \\
    \cline{1-7}
    MonoDDE (ours) & \underline{11.13} & \underline{7.32} & \underline{6.67} & \underline{5.94} & \textbf{3.78} & \textbf{3.33} \\
    \hline \hline
    \end{tabular}}
    \caption{Performance comparison on the Pedestrian and Cyclist classes of KITTI.} \label{Performance contrast on Pedestrian and Cyclist}
    \vspace{-0.3cm}
\end{table}

As shown in Table \ref{Performance contrast with SOTAs on Car}, taking monocular images as input, MonoDDE outperforms all other methods by large margins on both the testing and validation sets without introducing any extra information. For instance, MonoDDE surpasses Autoshape, a very recent SOTA method that utilizes CAD models as an extra clue, by 2.97\% for ${\rm AP}_{3D}70$ on the Moderate level. In other words, MonoDDE outperforms AutoShape by 20.96\% (2.97$\div$14.17) relatively.

In Table \ref{Performance contrast on Pedestrian and Cyclist}, MonoDDE outperforms all sparse-depth methods (M3D-RPN, MonoPair, DFR-Net and MonoFlex) significantly. Although MonoDDE is slightly weaker than CaDDN (a pseudo-LiDAR method) for the Pedestrian class, MonoDDE is much faster (MonoDDE: 0.04s/image vs. CaDDN: 0.63s/image). We speculate that MonoDDE does not behave the best for Pedestrian because pedestrians are non-rigid and much smaller compared with cars. Therefore, it is difficult to recognize the keypoints of pedestrians, while the pseudo-LiDAR methods do not suffer from this issue. 

\ParaSpace
\subsection{Ablation Study on Depth Estimation} \label{Study on Depth Estimation Methods}
\ParaSpace

This subsection aims to study how various depth estimation methods affect the 3D object detection precision. To this end, we compare the performance of the model that predict depths based on different combinations of three strategies (\textit{direct depth estimation}, \textit{depth from height} and \textit{depth from keypoint}). The model is trained on the KITTI training set and evaluated on the Car class of KITTI validation set. The results are reported in Table \ref{Ablation study on depth estimation}.

\begin{table}[htbp] 
    \vspace{-0.2cm}
    \centering
    \resizebox{82mm}{16mm}{
    \begin{tabular}{c|c|c|ccc|ccc}
    \hline \hline
    \multirow{2}{*}{E} & \multirow{2}{*}{H} & \multirow{2}{*}{K} & \multicolumn{3}{c|}{Val, ${\rm AP}_{3D}70$ (\%)} & \multicolumn{3}{c}{Val, ${\rm AP}_{BEV}70$ (\%)} \\
    \cline{4-9}
    & & & Easy & Moderate & Hard & Easy & Moderate & Hard\\
    \cline{1-9}
    \checkmark & & & 24.20 & 18.01 & 15.88 & 32.53 & 24.52 & 21.33 \\
    & \checkmark & & 25.01 & 18.36 & 15.32 & 33.15 & 24.83 & 21.40 \\
    & & \checkmark & 24.48 & 18.74 & 15.88 & 32.89 & 25.29 & 21.51 \\
    \cline{1-9}
    \checkmark & \checkmark & & 25.26 & 18.74 & 16.26 & 33.68 & 25.26 & 21.95 \\
    & \checkmark & \checkmark & 24.48 & 18.82 & 15.96 & 33.69 & 25.47 & 22.22 \\
    \checkmark & & \checkmark & 25.64 & 19.18 & 16.29 & 34.14 & 25.65 & 22.43 \\
    \cline{1-9}
    \rowcolor{Light} \checkmark & \checkmark & \checkmark & 26.66 & 19.75 & 16.72 & 35.51 & 26.48 & 23.07 \\
    \hline \hline
    \end{tabular}}
    \caption{Ablation study on depth estimation strategies (E: \textit{direct depth estimation}, H: \textit{depth from height}, and K: \textit{depth from keypoint}). We highlight the strategy adopted by MonoDDE in \colorbox{Light}{pink}.} \label{Ablation study on depth estimation}
    \vspace{-0.3cm}
\end{table}

As reported in the 1st--3rd rows of results in Table \ref{Ablation study on depth estimation},  when the three depth estimation strategies are applied separately, the model based on \textit{depth from keypoint} achieves the best performance on Moderate and Hard, and the one with only \textit{direct depth estimation} performs the worst. The underlying reason is that \textit{depth from keypoint} brings the most clues (16 depths) for every target while \textit{direct depth estimation} only produces 1 depth. %To verify this issue, we further evaluate the performance of the model that solves depth using the height information of center vertical line $H_{5}$, which is illustrated in Figure \ref{Depth solving system}. Staying consistent with \textit{Direct depth estimation}, it generates 1 estimation for every target. The obtained ${\rm AP}_{3D}70$ on KITTI validation set for the Car class are 24.31\%, 18.08\%, 15.21\% for the three difficulty levels, which are similar to the results of \textit{Direct depth estimation}.

According to the results in the last 4 rows of Table \ref{Ablation study on depth estimation}, if we combine two of the depth solving strategies to produce depths, better results are obtained because the diversity of the estimations is enhanced.  The best performance is achieved when we combine all the three strategies, which totally generates 20 depths for every detected object.

\begin{figure*}[htbp]
    \centering
    \includegraphics[scale=0.5]{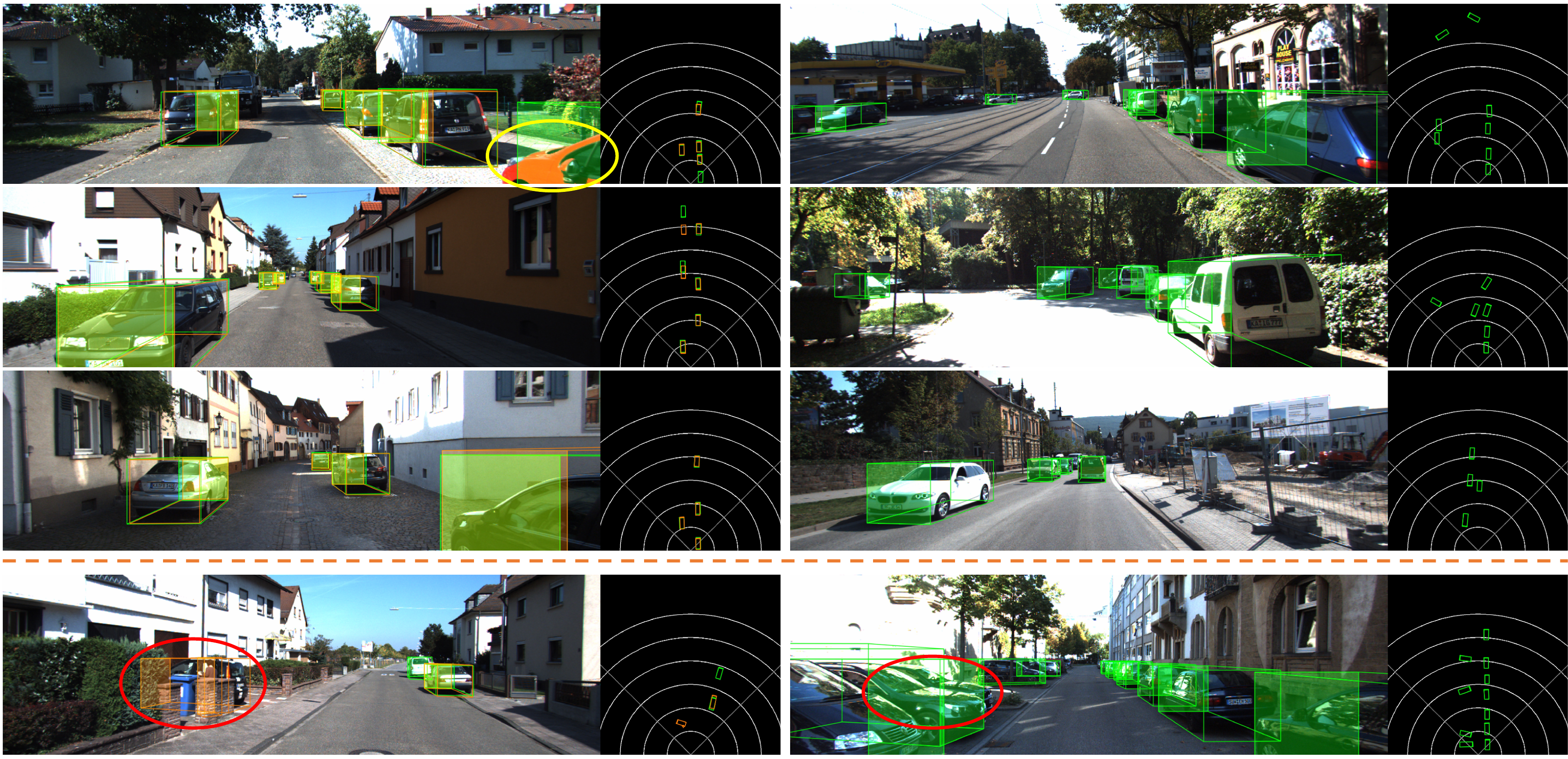}
    \caption{Visualization of some examples on the validation (left) and test (right) sets of KITTI. Failure cases are presented in the last row. The 3D green boxes are produced by MonoDDE and the orange boxes are the ground truth. We hightlight the target failed to be labeled by the annotator with a yellow ellipse and the undetected targets with red ellipses.} \label{Qualitative visualization}
    \vspace{-0.4cm}
\end{figure*}

\begin{table}[tbp] 
    \vspace{-0.2cm}
    \centering
    \resizebox{82mm}{14mm}{
    \begin{tabular}{c|c|ccc|ccc}
    \hline \hline
    \multirow{2}{*}{Select} & \multirow{2}{*}{Combine} & \multicolumn{3}{c|}{Val, ${\rm AP}_{3D}70$ (\%)} & \multicolumn{3}{c}{Val, ${\rm AP}_{BEV}70$ (\%)} \\
    \cline{3-8}
    & & Easy & Moderate & Hard & Easy & Moderate & Hard \\
    \cline{1-8}
    None & Hard & 25.71 & 19.13 & 16.39 & 34.30 & 25.72 & 22.39 \\
    None & Mean & 18.08 & 14.31 & 12.34 & 24.60 & 19.10 & 16.71 \\
    None & Weighted & 25.81 & 19.26 & 16.34 & 34.25 & 25.83 & 22.50 \\
    Min & Weighted & 26.31 & 19.59 & 16.58 & 34.79 & 26.09 & 22.78 \\
    \rowcolor{Light} Iterative & Weighted & 26.66 & 19.75 & 16.72 & 35.51 & 26.48 & 23.07 \\
    \cline{1-8}
    \rowcolor{Gray} Oracle & None & 49.96 & 38.73 & 33.06 & 58.69 & 43.96 & 37.65 \\
    \hline \hline
    \end{tabular}}
    \caption{Analysis of the depth selection and combination strategies.} \label{Analysis on depth selection and combination strategies}
    \vspace{-0.3cm}
\end{table}

\ParaSpace
\subsection{Analysis on Depth Selection and Combination} \label{Study on Depth Selection Strategies}
\ParaSpace

In this subsection, we analyze how the depth selection and combination strategies affect the results. We compare the performance of the model that tackles estimations in various ways. The results are presented in Table \ref{Analysis on depth selection and combination strategies}. The 1st column indicates how reliable depths are selected. Specifically, ``None" means no selection is applied. ``Min" indicates that the minimum estimated variance is regarded as the variance of the set $S$. ``Iterative" refers to the proposed iterative strategy described in Algorithm \ref{Robust depth combination pseudo code}. In the 2nd column, ``Hard" denotes that we use the value with the minimum variance as the combined depth $z_{c}$. ``Mean" and ``Weighted" represent that $z_{c}$ is the mean and the weighted sum of the depth estimations, respectively. Notably, the last row (in \colorbox{Gray}{gray}) of Table \ref{Analysis on depth selection and combination strategies} shows the performance if the best one is always selected from the set of the 20 depths. The strategy employed by MonoDDE is highlighted in \colorbox{Light}{pink}.

Comparing the 2nd and 3rd rows of the results in Table \ref{Analysis on depth selection and combination strategies}, we can notice that it is necessary to model the variance of the network output and combine estimations with the weighted sum operation in Eq.~(\ref{Eq19}). Besides, according to the 3rd and 5th rows of Table \ref{Analysis on depth selection and combination strategies}, removing outliers with Algorithm \ref{Robust depth combination pseudo code} boosts the detection precision effectively.

Notably, as presented in the last row of Table \ref{Analysis on depth selection and combination strategies}, if we develop a perfect strategy that always selects the most accurate one from the 20 depths, the ${\rm AP}_{3D}70$ on the Moderate level arrives 38.73\%. This phenomenon indicates how to select accurate depths deserves further study in the future work. 

\ParaSpace
\subsection{Analysis on the 3D Geometry Confidence} \label{Study on MD Confidence}
\ParaSpace
In this subsection, we study how various ways of modeling ${\rm P}_{3d|2d}$ affect the performance of MonoDDE. We compare the models based on different strategies, and the results are presented in Table \ref{Analysis on modeling 3D confidence strategies}. In the 1st column of Table \ref{Analysis on modeling 3D confidence strategies}, ``None" means we directly regard the 2D categorical heatmap score as the detection confidence ${\rm P}_{m}$. For ``3D IOU'', we train a specific network head to regress ${\rm P}_{3d|2d}$ defined based on 3D IOU. Denoting the 3D IOU between an estimated box and its ground truth as ${\rm I}_{3D}$, ${\rm P}_{3d|2d} = \min\{\max\{2 {\rm I}_{3D} - 0.5, 0\}, 1\}$ following \cite{li2021monocular}. ``$d_{1}$--$d_{20}$" indicates that ${\rm P}_{3d|2d}$ is computed based on the confidences of the 20 depth estimations through the weighted sum ($\sum\limits_{i=1}^{20}\omega_{i}d_{i}$) like Eq.~(\ref{Eq23}). ``$d_{c}$" and ``$d_{b}$" mean we model ${\rm P}_{3d|2d}$ using the combined depth confidence $d_{c}$ and the 3D box confidence $d_{b}$, respectively. ``3D Confidence" is the strategy employed by MonoDDE (marked in \colorbox{Light}{pink}).

\begin{table}[tbp] 
    \vspace{-0.2cm}
    \centering
    \resizebox{82mm}{14mm}{
    \begin{tabular}{c|ccc|ccc}
    \hline \hline
    \multirow{2}{*}{Strategies} & \multicolumn{3}{c|}{Val, ${\rm AP}_{3D}70$ (\%)} & \multicolumn{3}{c}{Val, ${\rm AP}_{BEV}70$ (\%)} \\
    \cline{2-7}
    & Easy & Moderate & Hard & Easy & Moderate & Hard \\
    \cline{1-7}
    None & 23.67 & 18.15 & 15.41 & 31.59 & 24.57 & 21.45 \\
    3D IOU & 22.67 & 18.54 & 16.06 & 30.30 & 24.14 & 21.17 \\
    $d_{1}$--$d_{20}$ & 25.32 & 19.08 & 16.12 & 33.37 & 25.39 & 22.16 \\
    $d_{c}$ & 25.58 & 19.12 & 16.17 & 33.76 & 25.72 & 22.34 \\
    $d_{b}$ & 26.02 & 19.48 & 16.43 & 34.14 & 25.87 & 22.88 \\
    \rowcolor{Light} 3D Confidence & 26.66 & 19.75 & 16.72 & 35.51 & 26.48 & 23.07 \\
    \hline \hline
    \end{tabular}}
    \caption{Analysis of modeling conditional 3D confidence strategies. } \label{Analysis on modeling 3D confidence strategies}
    \vspace{-0.5cm}
\end{table}

From Table \ref{Analysis on modeling 3D confidence strategies}, we can mainly observe two facts: (1) Comparing the 1st and 2nd rows of the results, it can be found that modeling ${\rm P}_{3d|2d}$ with 3D IOU does not always boost the performance. (2) According to the values in the 3rd--6th rows, modeling ${\rm P}_{3d|2d}$ with our proposed strategy leads to the best results.

\ParaSpace
\subsection{Qualitative Results and Limitation} \label{Qualitative Results}
\ParaSpace
We show some 3D boxes and BEV maps produced by MonoDDE on both the KITTI validation and testing sets in Figure \ref{Qualitative visualization}. As shown, although some targets are not labeled by the annotators, they are still detected by MonoDDE correctly. However, as illustrated in the last row of Figure \ref{Qualitative visualization}, similar to other works, the performance of MonoDDE on detecting seriously occluded targets is limited.

\section{Conclusion}
In this paper, we have proposed a robust monocular 3D detector that can produce diverse depth estimations for every target and combine the reliable estimations into a single depth. Besides, a new way for modeling the conditional 3D confidence is developed. The experimental results indicate that all our proposed techniques are effective, which establish new SOTA in monocular 3D object detection. We hope this work can shed light on how to tackle the problem of missing depth information in monocular images. We thank MindSpore \cite{MindSpore} for the partial support to this work, which is a new deep learning computing framework.

%%%%%%%%% REFERENCES
{\small
\bibliographystyle{ieee_fullname}
\bibliography{reference}
}

\end{document}